\documentclass[a4paper,twoside]{article}

\usepackage{epsfig}
\usepackage{subcaption}
\usepackage{calc}
\usepackage{amssymb}
\usepackage{amstext}
\usepackage{amsmath}
\usepackage{amsthm}
\usepackage{multicol}
\usepackage{pslatex}
\usepackage{apalike}
\usepackage{algorithm2e}
\usepackage[bottom]{footmisc}

\usepackage[inline]{enumitem}
\usepackage[hidelinks]{hyperref}
\usepackage{tikz}
\usetikzlibrary{shapes.geometric, positioning}
\usepackage{xspace}
\newcommand*{\ie}{i.e.\@\xspace}
\newcommand*{\eg}{e.g.\@\xspace}
\newcommand*{\wrt}{w.r.t.\@\xspace}
\usepackage{pifont}
\newcommand{\cmark}{\text{\ding{51}}}
\newcommand{\xmark}{\text{\ding{55}}}
\usepackage{datetime}
\newdate{date}{12}{09}{2024}

\usepackage{SCITEPRESS}     

\definecolor{my-light-blue}{RGB}{218,232,252}
\definecolor{my-light-green}{RGB}{213,232,212}
\definecolor{my-light-orange}{RGB}{255,230,204}

\def \globalscale {1.000000}
\pgfmathsetmacro{\pegheight}{0.6} 
\pgfmathsetmacro{\tablewidth}{1.0} 
\pgfmathsetmacro{\tableheight}{0.6} 
\pgfmathsetmacro{\tableslotheight}{0.25} 
\pgfmathsetmacro{\tablepenetrationdepth}{0.25} 
\pgfmathsetmacro{\gripperwidth}{0.5} 
\pgfmathsetmacro{\gripperfingerwidth}{0.125} 
\pgfmathsetmacro{\gripperheigth}{0.1875} 
\pgfmathsetmacro{\gripperfingerbaselength}{0.25} 
\pgfmathsetmacro{\gripperfingertiplength}{0.125} 

\pgfmathsetmacro{\gripperbodywidth}{\gripperwidth - 2 * \gripperfingerwidth}
\pgfmathsetmacro{\gripperxi}{0.5 * \tablewidth - 0.5 * \gripperbodywidth - \gripperfingerwidth}
\pgfmathsetmacro{\gripperxf}{\gripperxi + \gripperwidth}
\pgfmathsetmacro{\tablesidewidth}{0.5 * (\tablewidth - \gripperbodywidth)}
\pgfmathsetmacro{\pegwidth}{\gripperbodywidth}
\pgfmathsetmacro{\pegcenterx}{0.5 * \tablewidth}
\pgfmathsetmacro{\pegcentery}{0}
\pgfmathsetmacro{\pegxi}{\pegcenterx - 0.5 * \pegwidth}
\pgfmathsetmacro{\pegxf}{\pegcenterx + 0.5 * \pegwidth}
\pgfmathsetmacro{\pegyi}{\pegcentery - 0.5 * \pegheight}
\pgfmathsetmacro{\pegyf}{\pegcentery + 0.5 * \pegheight}
\pgfmathsetmacro{\tableslotcenterx}{0.5 * \tablewidth}
\pgfmathsetmacro{\tableslotcentery}{\tableslotheight + 0.5 * \tablepenetrationdepth}
\pgfmathsetmacro{\textx}{\tablewidth - 0.5 * \tablesidewidth}
\pgfmathsetmacro{\texty}{\pegcentery}

\pgfmathsetmacro{\gripperyf}{\pegyf + \gripperheigth}
\pgfmathsetmacro{\gripperbaseyi}{\gripperyf - \gripperheigth}
\pgfmathsetmacro{\gripperfingertipyf}{\gripperyf - \gripperfingerbaselength}
\pgfmathsetmacro{\gripperfingertipyi}{\gripperfingertipyf - \gripperfingertiplength}

\begin{document}

\title{
On the role of Artificial Intelligence methods in modern force-controlled manufacturing robotic tasks
}

\author{\authorname{Vincenzo Petrone\orcidAuthor{0000-0003-4777-1761},
Enrico Ferrentino\orcidAuthor{0000-0003-0768-8541
} and
Pasquale Chiacchio\orcidAuthor{0000-0003-3385-8866}}
\affiliation{Department of Information Engineering, Electrical Engineering and Applied Mathematics (DIEM), University of Salerno, 84084 Fisciano, Italy}
\email{\{vipetrone, eferrentino, pchiacchio\}@unisa.it}
\vspace*{-4mm}\affiliation{This work has been accepted for publication at ICINCO 2024 - 21st International Conference on Informatics in Control, Automation and Robotics on \displaydate{date}.\\
Science and Technology Publications, Lda holds the copyright on the published version of this article.\\
Please refer to the published version in the conference proceedings at \url{https://www.doi.org/10.5220/0013013300003822}}
\vspace*{-4mm}
}


\abstract{
This position paper explores the integration of Artificial Intelligence (AI) into force-controlled robotic tasks within the scope of advanced manufacturing, a cornerstone of Industry 4.0.
AI's role in enhancing robotic manipulators -- key drivers in the Fourth Industrial Revolution -- is rapidly leading to significant innovations in smart manufacturing.
The objective of this article is to frame these innovations in practical force-controlled applications -- \eg deburring, polishing, and assembly tasks like peg-in-hole (PiH) -- highlighting their necessity for maintaining high-quality production standards.
By reporting on recent AI-based methodologies, this article contrasts them and identifies current challenges to be addressed in future research.
The analysis concludes with a perspective on future research directions, emphasizing the need for common performance metrics to validate AI techniques, integration of various enhancements for performance optimization, and the importance of validating them in relevant scenarios.
These future directions aim to provide consistency with already adopted approaches, so as to be compatible with manufacturing standards, increasing the relevance of AI-driven methods in both academic and industrial contexts.
}

\onecolumn \maketitle \normalsize \setcounter{footnote}{0} \vfill

\section{\uppercase{Introduction}}\label{sec:introduction}

Manufacturing processes are nowadays experiencing the peak of their technological evolution, spurred by rapid advances in industrialization methods currently developing in the ongoing Fourth Industrial Revolution \cite{xu_industry_2018}.
These emerging innovations are defining the next generation of industries, leading to the so-called \textit{Industry 4.0} \cite{yang_industry_2021}.

One of the pillars of the revolution is Artificial Intelligence (AI), whose application has seen tremendous advancements and increasing popularity in recent years.
Robotic manipulators, which were already one of the core drivers of the Third Industrial Revolution, are now amongst the technologies that are benefiting the most from AI \cite{bai_industry_2020}.
Merging these two powerful technologies is leading to the rise of \textit{advanced manufacturing} (also termed \textit{smart manufacturing}), which constitutes the foundation of Industry 4.0 \cite{yang_industry_2021}.

This position paper presents a discussion on a peculiar sector of robotic tasks, namely \textit{force-controlled} applications.
Such tasks are of fundamental practical importance in manufacturing, since they deal with manipulators exerting forces on the working environment, with the objective of, \eg, refining a workpiece or manipulating and assembling objects.

Popular force-controlled tasks encompass, for instance, deburring \cite{lloyd_precision_2024}, polishing \cite{iskandar_hybrid_2023}, and assembly \cite{luo_reinforcement_2019}.
On the one hand, the first two require exerting a specific normal force on the working surface, in order to remove burrs and production inaccuracies of a workpiece, or to smooth and finish the surface itself.
On the other hand, the latter consists in the assembly of two or more objects together: the most renowed example in this context is \textit{peg-in-hole} (PiH) \cite{sorensen_online_2016}.
Although different in terms of requirements, all of the aforementioned tasks necessitate to control, directly or indirectly, the forces exchanged between the manipulator and the working environment.

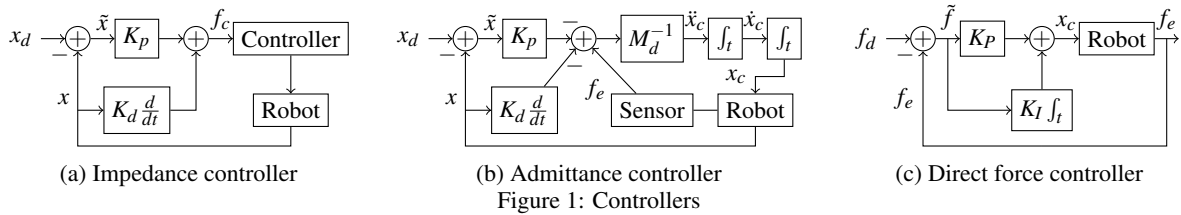
\begin{figure*}[t]
    \centering

    \begin{subfigure}{0.3\textwidth}
        \centering
        \begin{tikzpicture}[auto, node distance=1em]
    \newcommand*{\schemefontsize}{\footnotesize}
    
    \schemefontsize
    
    \tikzstyle{block} = [draw, rectangle, minimum height=1em, minimum width=1em, font=\schemefontsize]
    \tikzstyle{sum} = [draw, circle, inner sep=0pt, minimum size=1em, font=\schemefontsize]
    \newcommand*{\verticaldistance}{3em}

    \node at (0, 0) (xd) {$x_d$};  
    \node [sum, right=of xd] (error) {$+$};  
    \node [block, right=of error] (proportional) {$K_p$};  
    \node [block, below of=proportional, node distance=\verticaldistance] (derivative) {$K_d\frac{d}{dt}$};  
    \node [sum, right=of proportional] (sum) {$+$};  
    \node [block, right=of sum] (controller) {Controller};  
    \node [block, below of=controller, node distance=\verticaldistance] (robot) {Robot};  
    \node [coordinate, below of=robot, node distance=1.5em] (x) {};  

    \draw [->] (xd) -- (error);  
    \draw [->] (robot) -- (x) -| node[near end] {$x$} node[pos=0.98] {$-$} (error);  
    \draw [->] (error) -- node {$\tilde x$} (proportional);  
    \draw [->] (proportional) -- (sum);  
    \draw [->] (derivative) -| (sum);  
    \draw [->] (sum) -- node {$f_c$} (controller);  
    \draw [->] (controller) -- (robot);  
    \node [coordinate, below of=error, node distance=\verticaldistance] (aux) {};
    \draw [->] (aux) -- (derivative); 
\end{tikzpicture}
        \caption{Impedance controller}
        \label{fig:impedance-controller}
    \end{subfigure}
    \hfill
    \begin{subfigure}{0.35\textwidth}
        \centering
        \begin{tikzpicture}[auto, node distance=1em]
    \newcommand*{\schemefontsize}{\footnotesize}
    
    \schemefontsize
    
    \tikzstyle{block} = [draw, rectangle, minimum height=1em, minimum width=1em, font=\schemefontsize]
    \tikzstyle{sum} = [draw, circle, inner sep=0pt, minimum size=1em, font=\schemefontsize]
    \newcommand*{\verticaldistance}{3em}    

    \node at (0, 0) (ref) {$x_d$};  
    \node [sum, right=of ref] (error) {$+$};  
    \node [block, right=of error] (proportional) {$K_p$};  
    \node [block, below of=proportional, node distance=\verticaldistance] (derivative) {$K_d\frac{d}{dt}$};  
    \node [sum, right=of proportional] (sum) {$+$};  
    \node[block, right=of sum] (mass) {$M_d^{-1}$};
    \node[block, right=of mass] (int1) {$\int_t$};
    \node[block, right=of int1] (int2) {$\int_t$};
    \node [block, below of=mass, node distance=\verticaldistance] (sensor) {Sensor};
    \node [coordinate, below of=sensor, node distance=1.5em] (feedback) {};
    \node [block, right=of sensor] (robot) {Robot};  
    
    \draw [->] (ref) -- (error);  
    \draw [->] (error) -- node {$\tilde x$} (proportional);  
    \draw [->] (proportional) -- node[pos=0.98] {$-$} (sum);  
    \draw [->] (derivative) -- node[pos=0.98, below] {$-$} (sum);  
    \draw [->] (sum) -- (mass);  
    \draw [->] (mass) -- node {$\ddot x_c$} (int1);  
    \draw [->] (int1) -- node {$\dot x_c$} (int2); 
    \path (int2) -- (robot) coordinate[pos=0.5] (aux);
    \draw [->] (int2) |- (aux) -| node[left] {$x_c$} (robot); 
    
    \draw [-] (robot) -- (sensor);
    \draw [->] (sensor) -- node {$f_e$} (sum);
    \draw [->] (robot) |- (feedback) -| node[pos=0.98] {$-$} node[near end] {$x$} (error);  
    \node [coordinate, below of=error, node distance=\verticaldistance] (x) {};
    \draw [->] (x) -- (derivative);

\end{tikzpicture}
        \caption{Admittance controller}
        \label{fig:admittance-controller}
    \end{subfigure}
    \hfill
    \begin{subfigure}{0.3\textwidth}
        \centering
        \begin{tikzpicture}[auto, node distance=1em]
    \newcommand*{\schemefontsize}{\footnotesize}
    
    \schemefontsize
    
    \tikzstyle{block} = [draw, rectangle, minimum height=1em, minimum width=1em, font=\schemefontsize]
    \tikzstyle{sum} = [draw, circle, inner sep=0pt, minimum size=1em, font=\schemefontsize]
    \newcommand*{\verticaldistance}{3em}

    \node at (0, 0) (ref) {$f_d$};  
    \node [sum, right=of ref] (error) {$+$};  
    \node [block, right=of error] (proportional) {$K_P$};  
    \node [sum, right=of proportional] (sum) {$+$};  
    \node [block, below of=sum, node distance=\verticaldistance] (integral) {$K_I\int_t$};  
    \node [block, right=of sum] (robot) {Robot};  
    \node [coordinate, right=of robot] (output) {};  
    \node [coordinate, below of=integral, node distance=1.5em] (sensor) {};  
    
    \draw [->] (ref) -- (error);  
    \draw [->] (error) -- node[name=aux] {$\tilde f$} (proportional);  
    \draw [->] (aux) |- (integral);  
    \draw [->] (proportional) -- (sum);  
    \draw [->] (integral) -- (sum);  
    \draw [->] (sum) -- node {$x_c$} (robot);  
    \draw [->] (robot) -- node[name=y] {$f_e$} (output);  
    
    \draw [-] (y) |- (sensor);  
    \draw [->] (sensor) -| node[pos=0.98] {$-$} node[near end] {$f_e$} (error);  

\end{tikzpicture}
        \caption{Direct force controller}
        \label{fig:direct-force-controller}
    \end{subfigure}
    
    \caption{Controllers}
    \label{fig:controller}
\end{figure*}

This article discusses some modern techniques proposed by recent literature, focusing on their AI-based counterparts, elaborating on how they tackle the practical issues arising from the aforementioned force-controlled applications, and highlighting differences among them.
To better frame the scope of this article, we stress that we are not proposing a formal and extensive literature review, but we are discussing recent relevant research to claim our position on what the current challenges are in the aforementioned contexts, and propose possible future research directions.

The rest of this paper is organized as follows.
Section~\ref{sec:motivation} motivates the need of AI in force-controlled tasks, and presents in detail typical applications in such contexts.
Section~\ref{sec:challenges} elaborates on the challenges AI-based methods are currently facing, comparing them with state-of-the-art baselines.
Section~\ref{sec:future-directions} summarizes the discussion and poses questions to guide future research.
Section~\ref{sec:conclusions} concludes the paper, stating the main findings of the conducted analysis.

\section{\uppercase{Motivation}}\label{sec:motivation}

\subsection{Preliminaries}

\begin{figure*}[t!]
    \centering

    \begin{subfigure}{0.24\textwidth}
        \centering
        \includegraphics[height=125pt]{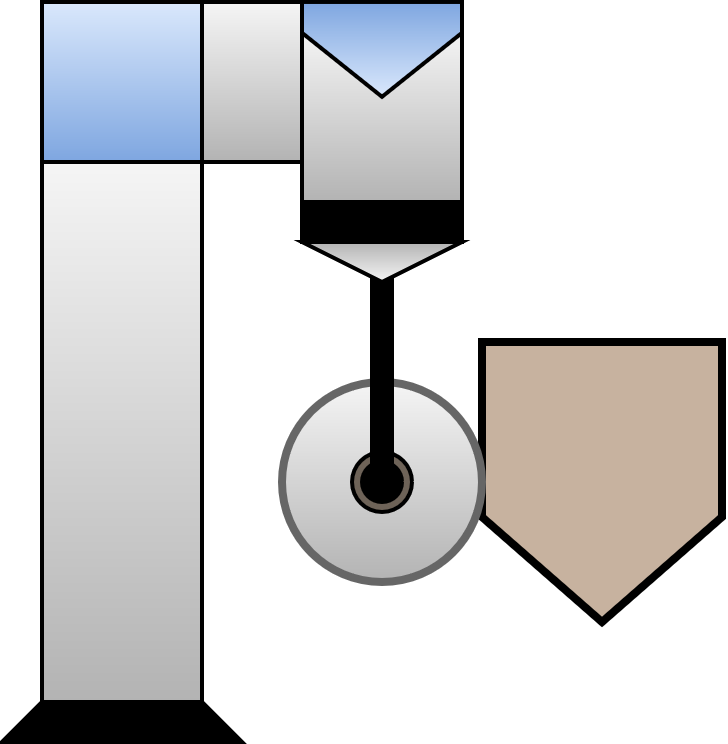}
        \caption{Deburring}
        \label{fig:deburring}
    \end{subfigure}
    \hfill
    \begin{subfigure}{0.48\textwidth}
        \centering
        \includegraphics[height=125pt]{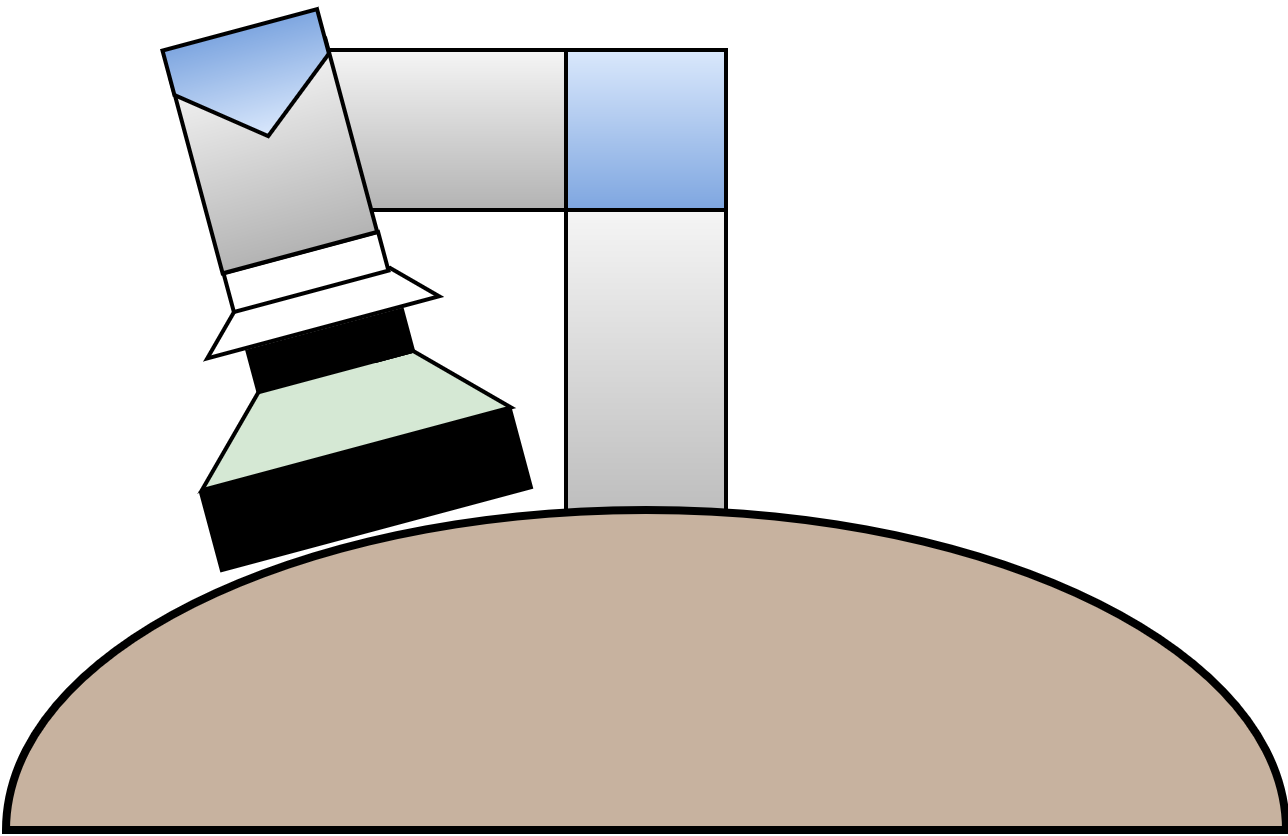}
        \caption{Polishing}
        \label{fig:polishing}
    \end{subfigure}
    \begin{subfigure}{0.24\textwidth}
        \centering
        \begin{tikzpicture}[y=1\textwidth, x=1\textwidth, xscale=\globalscale, yscale=\globalscale, every node/.append style={scale=\globalscale}, inner sep=0pt, outer sep=0pt]

    \pgfmathsetmacro{\tablepenetrationdepth}{0.075}
    \pgfmathsetmacro{\rotationangle}{0}
    \pgfmathsetmacro{\shiftx}{0}
    \pgfmathsetmacro{\shifty}{\tableslotheight + \tablepenetrationdepth + 0.5 * \pegheight}

    \begin{scope}[xshift=\shiftx\textwidth, yshift=\shifty\textwidth]
        \node (peg) [
            draw=black,
            fill=my-light-blue,
            very thick,
            minimum width=\pegwidth\textwidth,
            minimum height=\pegheight\textwidth,
            anchor=center,
            rotate=\rotationangle] at (\pegcenterx, \pegcentery) {};

        \begin{scope}[rotate around={\rotationangle: (peg.center)}]
            \path[draw=black, fill=my-light-green, very thick, miter limit=10.0]
                (\gripperxi, \gripperyf) --
                (\gripperxf, \gripperyf) --
                (\gripperxf, \gripperfingertipyf) --
                (\pegxf,\gripperfingertipyi) --
                (\pegxf, \gripperbaseyi) --
                (\pegxi, \gripperbaseyi) --
                (\pegxi, \gripperfingertipyi) --
                (\gripperxi, \gripperfingertipyf) --
                cycle;
            \path[draw=black, miter limit=10.0]
                (\pegxi, \gripperbaseyi) --
                (\pegxi, \gripperyf);
            \path[draw=black, miter limit=10.0]
                (\pegxf, \gripperbaseyi) --
                (\pegxf, \gripperyf);
        \end{scope}
    \end{scope}
    
    \path[draw=black,fill=my-light-orange,very thick,miter limit=10.0]
        (\tablesidewidth, \tableheight) --
        (\tablesidewidth, \tableslotheight) --
        (\tablewidth-\tablesidewidth, \tableslotheight) --
        (\tablewidth-\tablesidewidth, \tableheight) --
        (\tablewidth, \tableheight) --
        (\tablewidth, 0) --
        (0, 0) --
        (0, \tableheight) --
        cycle;
\end{tikzpicture}
\vfill
        \caption{Peg-in-hole}
        \label{fig:peg-in-hole}
    \end{subfigure}
    
    \caption{Force-controlled tasks}
    \label{fig:force-controlled-tasks}
\end{figure*}
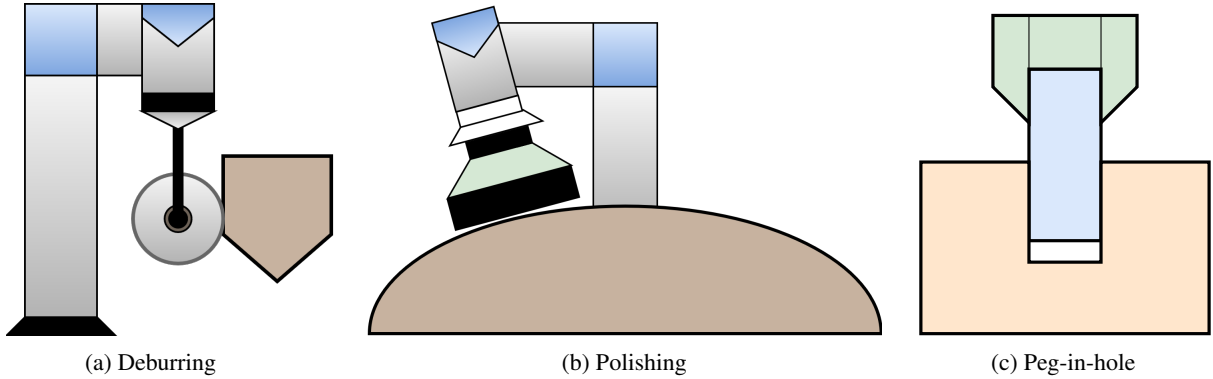

Force control is an essential requirement in a vast number of applications of utmost importance in a broad spectrum of real-world contexts, ranging from industrial \cite{lloyd_precision_2024} to medical \cite{tang_meta-learning-based_2023} scenarios.
For this reason, force control has been one of the major interests in robotics research in the last decades.

With the fundamental objectives of ensuring safety and preserving environment integrity, indirect force control methods, \eg impedance \cite{hogan_impedance_1985} and admittance \cite{newman_stability_1992} controllers, have been proposed to regulate the manipulator behavior, generalizing pure motion control to interaction scenarios and providing robots with compliant characteristics with respect to the environment they are in contact with.

Usually, an impedance control law (Figure~\ref{fig:impedance-controller}) can be formulated as
\begin{equation}\label{eq:impedance-controller}
    f_c = K_p \tilde x + K_d \dot x,
\end{equation}
where $K_p$ and $K_d$ are stiffness and damping parameters, $\tilde x \triangleq x_d - x$ is the task-space error between the setpoint $x_d$ and the actual end-effector (EE) position $x$, and $f_c$ is a Cartesian wrench to control via a torque-based controller, which also compensates for the manipulator's internal dynamics, in order to impose a compliant interaction at the EE.
On the other hand, admittance control (Figure~\ref{fig:admittance-controller}) takes the form
\begin{equation}\label{eq:admittance-controller}
    \ddot x_c = M_d^{-1} \left( f_e - K_d \dot x - K_p \tilde x \right),
\end{equation}
where $M_d$ is the mass matrix of virtual mass-spring-damper dynamics EE wrenches $f_e$ (usually measured with a force/torque sensor, mounted at the manipulator's flange) are input to.
The resulting task-space acceleration $\ddot x_c$ is usually commanded to a low-level motion controller.

As an additional performance requirement, industrial tasks usually demand a desired force to be accurately exerted on the working surface: in this case, direct force control (DFC) strategies can be employed to track a reference force \cite{khatib_unified_1987}.
Typical examples of force-related tasks are illustrated in Figure~\ref{fig:force-controlled-tasks}, \ie workpiece deburring (Figure~\ref{fig:deburring}) and surface polishing (Figure~\ref{fig:polishing}).
Typically, this is implemented with a PI loop as
\begin{equation}\label{eq:direct-force-control}
    x_c = K_P \tilde f + K_I \int_t{ \tilde f dt},
\end{equation}
where $K_P$ and $K_I$ are proportional and integral control parameters, respectively, and $\tilde f \triangleq f_d - f_e$ is the wrench error between the desired wrench $f_d$ and the actual one (Figure~\ref{fig:direct-force-controller}).

\subsection{AI-based methods}

One of the most challenging issues force control algorithms have to face is the inaccuracy and unpredictability of the environment geometry and dynamics.
Most of advanced techniques are based on the assumption that the environment force follows a simple linear spring model in the form
\begin{equation}\label{eq:force-spring-model}
    f_e = K_e (x - x_r),
\end{equation}
with $K_e$ and $x_r$ being the (unknown) environment stiffness and rest position, respectively.
However, this assumption does not always hold, especially at high speeds \cite{iskandar_hybrid_2023}, and it represents, in general, an approximation, since it is not expected for the environment to behave linearly \cite{matschek_safe_2023}.
This aspect motivates the necessity of employing data-driven strategies, exploiting AI and Machine Learning (ML) to compensate for the inherent difficulties risen from the aforementioned complex scenarios.

Lack of accurate and reliable environment modeling becomes even more problematic in contact-rich assembly tasks, \eg, PiH (Figure~\ref{fig:peg-in-hole}) or similar dual setups, such as gear assembly \cite{luo_reinforcement_2019}.
These challenging scenarios suffer from additional problems, namely unknown or inaccurate locations of the objects to manipulate (for instance, the hole location or the held peg orientation in PiH), and the complexity of the forces due to frequent contacts, occurring because of the limited tolerance between the parts to assemble.
Powerful tools to cope with these inescapable uncertainties and complexities encompass vision-based approaches \cite{zhang_vision-based_2023}, to estimate hole locations, and Reinforcement Learning (RL), through which control policies are devised from experience and computed according to an objective function to maximize, usually modeling the task goal \cite{elguea-aguinaco_review_2023}.

A comprehensive review on PiH control strategies is available in \cite{jiang_state---art_2020}.
However, this survey does not consider some recent popular methods: indeed, the considered AI- and ML-based methods mostly rely on \textit{learning from demonstrations}, requiring the physical presence of a human operator in the loop to collect data and devise policies to deploy on the robot \cite{zhang_learning_2021}.

The next section will present some of the recent advancements in AI for force-related and contact-rich tasks, and will highlight the practical challenges they are demanded to tackle.

\section{\uppercase{Challenges}}\label{sec:challenges}

\subsection{Problem definition}\label{sec:problem-definition}

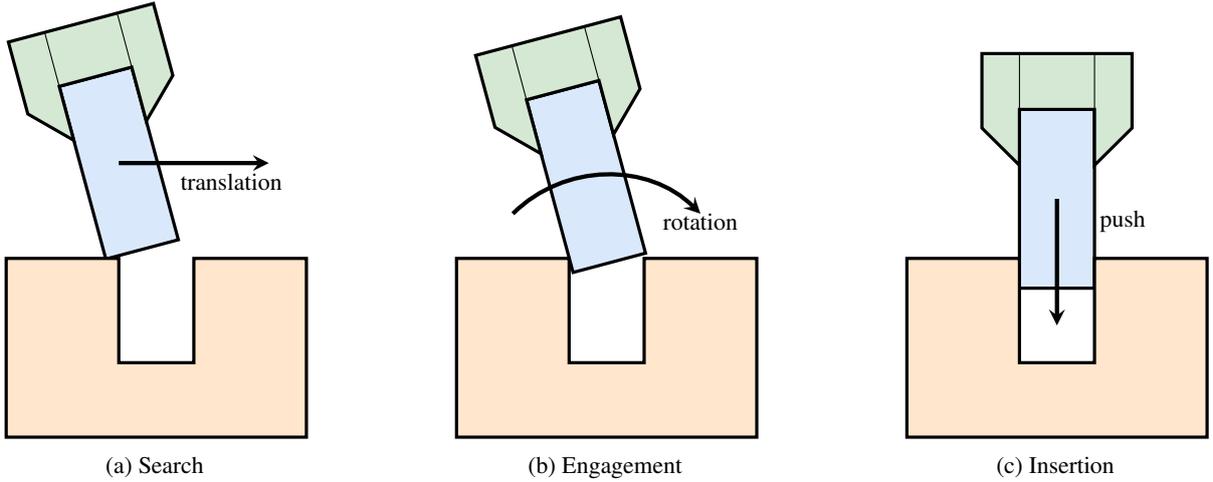
\begin{figure*}[t!]
    \centering

    \begin{subfigure}{0.25\textwidth}
        \centering
        \begin{tikzpicture}[y=1\textwidth, x=1\textwidth, xscale=\globalscale, yscale=\globalscale, every node/.append style={scale=\globalscale}, inner sep=0pt, outer sep=0pt]

    \pgfmathsetmacro{\rotationangle}{15}
    \pgfmathsetmacro{\shiftx}{-0.5 * \pegwidth}
    \pgfmathsetmacro{\shifty}{\tableheight + 0.5 * \pegheight + 0.02}

    \begin{scope}[xshift=\shiftx\textwidth, yshift=\shifty\textwidth]
        \node (peg) [
            draw=black,
            fill=my-light-blue,
            very thick,
            minimum width=\pegwidth\textwidth,
            minimum height=\pegheight\textwidth,
            anchor=center,
            rotate=\rotationangle] at (\pegcenterx, \pegcentery) {};

        \begin{scope}[rotate around={\rotationangle: (peg.center)}]
            \path[draw=black, fill=my-light-green, very thick, miter limit=10.0]
                (\gripperxi, \gripperyf) --
                (\gripperxf, \gripperyf) --
                (\gripperxf, \gripperfingertipyf) --
                (\pegxf,\gripperfingertipyi) --
                (\pegxf, \gripperbaseyi) --
                (\pegxi, \gripperbaseyi) --
                (\pegxi, \gripperfingertipyi) --
                (\gripperxi, \gripperfingertipyf) --
                cycle;
            \path[draw=black, miter limit=10.0]
                (\pegxi, \gripperbaseyi) --
                (\pegxi, \gripperyf);
            \path[draw=black, miter limit=10.0]
                (\pegxf, \gripperbaseyi) --
                (\pegxf, \gripperyf);
        \end{scope}
    \end{scope}
    
    \draw[draw=black, ultra thick, miter limit=10.0, -stealth]
        (peg.center) --node[near end, below, yshift=-4pt, font=\small]{translation} ++(0.5*\tablewidth, 0);
    
    \path[draw=black,fill=my-light-orange,very thick,miter limit=10.0]
        (\tablesidewidth, \tableheight) --
        (\tablesidewidth, \tableslotheight) --
        (\tablewidth-\tablesidewidth, \tableslotheight) --
        (\tablewidth-\tablesidewidth, \tableheight) --
        (\tablewidth, \tableheight) --
        (\tablewidth, 0) --
        (0, 0) --
        (0, \tableheight) --
        cycle;
\end{tikzpicture}
\vfill
        \caption{Search}
        \label{fig:peg-in-hole-search}
    \end{subfigure}
    \hfill
    \begin{subfigure}{0.25\textwidth}
        \centering
        \begin{tikzpicture}[y=1\textwidth, x=1\textwidth, xscale=\globalscale, yscale=\globalscale, every node/.append style={scale=\globalscale}, inner sep=0pt, outer sep=0pt]

    \pgfmathsetmacro{\rotationangle}{15}
    \pgfmathsetmacro{\shiftx}{-0.275 * \pegwidth}
    \pgfmathsetmacro{\shifty}{\tableheight + 0.5 * \pegheight - 0.025}
    \pgfmathsetmacro{\engagementarrowbeginx}{0.5 * \tablesidewidth}
    \pgfmathsetmacro{\engagementarrowbeginy}{\tableheight + 0.25 * \pegheight}
    \pgfmathsetmacro{\engagementarrowendx}{\tablewidth - 0.5 * \tablesidewidth}
    \pgfmathsetmacro{\engagementarrowendy}{\engagementarrowbeginy}
    \pgfmathsetmacro{\engagementarrowrotout}{45}
    \pgfmathsetmacro{\engagementarrowrotin}{180 - \engagementarrowrotout}

    \begin{scope}[xshift=\shiftx\textwidth, yshift=\shifty\textwidth]
        \node (peg) [
            draw=black,
            fill=my-light-blue,
            very thick,
            minimum width=\pegwidth\textwidth,
            minimum height=\pegheight\textwidth,
            anchor=center,
            rotate=\rotationangle] at (\pegcenterx, \pegcentery) {};

        \begin{scope}[rotate around={\rotationangle: (peg.center)}]
            \path[draw=black, fill=my-light-green, very thick, miter limit=10.0]
                (\gripperxi, \gripperyf) --
                (\gripperxf, \gripperyf) --
                (\gripperxf, \gripperfingertipyf) --
                (\pegxf,\gripperfingertipyi) --
                (\pegxf, \gripperbaseyi) --
                (\pegxi, \gripperbaseyi) --
                (\pegxi, \gripperfingertipyi) --
                (\gripperxi, \gripperfingertipyf) --
                cycle;
            \path[draw=black, miter limit=10.0]
                (\pegxi, \gripperbaseyi) --
                (\pegxi, \gripperyf);
            \path[draw=black, miter limit=10.0]
                (\pegxf, \gripperbaseyi) --
                (\pegxf, \gripperyf);
        \end{scope}
    \end{scope}
    
    \draw[draw=black, ultra thick, miter limit=10.0, -stealth]
        (\engagementarrowbeginx, \engagementarrowbeginy)
        to [out=\engagementarrowrotout, in=\engagementarrowrotin]
        node[below, pos=1.0, font=\small]{rotation}
        (\engagementarrowendx, \engagementarrowendy);
    
    \path[draw=black,fill=my-light-orange,very thick,miter limit=10.0]
        (\tablesidewidth, \tableheight) --
        (\tablesidewidth, \tableslotheight) --
        (\tablewidth-\tablesidewidth, \tableslotheight) --
        (\tablewidth-\tablesidewidth, \tableheight) --
        (\tablewidth, \tableheight) --
        (\tablewidth, 0) --
        (0, 0) --
        (0, \tableheight) --
        cycle;
\end{tikzpicture}
\vfill
        \caption{Engagement}
        \label{fig:peg-in-hole-engagement}
    \end{subfigure}
    \hfill
    \begin{subfigure}{0.25\textwidth}
        \centering
        \begin{tikzpicture}[y=1\textwidth, x=1\textwidth, xscale=\globalscale, yscale=\globalscale, every node/.append style={scale=\globalscale}, inner sep=0pt, outer sep=0pt]

    \pgfmathsetmacro{\rotationangle}{0}
    \pgfmathsetmacro{\shiftx}{0}
    \pgfmathsetmacro{\shifty}{\tableslotheight + \tablepenetrationdepth + 0.5 * \pegheight}
    \pgfmathsetmacro{\pusharrowend}{0.5 * \pegwidth + 0.5 * \tablesidewidth}

    \begin{scope}[xshift=\shiftx\textwidth, yshift=\shifty\textwidth]
        \node (peg) [
            draw=black,
            fill=my-light-blue,
            very thick,
            minimum width=\pegwidth\textwidth,
            minimum height=\pegheight\textwidth,
            anchor=center,
            rotate=\rotationangle] at (\pegcenterx, \pegcentery) {};

        \begin{scope}[rotate around={\rotationangle: (peg.center)}]
            \path[draw=black, fill=my-light-green, very thick, miter limit=10.0]
                (\gripperxi, \gripperyf) --
                (\gripperxf, \gripperyf) --
                (\gripperxf, \gripperfingertipyf) --
                (\pegxf,\gripperfingertipyi) --
                (\pegxf, \gripperbaseyi) --
                (\pegxi, \gripperbaseyi) --
                (\pegxi, \gripperfingertipyi) --
                (\gripperxi, \gripperfingertipyf) --
                cycle;
            \path[draw=black, miter limit=10.0]
                (\pegxi, \gripperbaseyi) --
                (\pegxi, \gripperyf);
            \path[draw=black, miter limit=10.0]
                (\pegxf, \gripperbaseyi) --
                (\pegxf, \gripperyf);
        \end{scope}
    \end{scope}
    
    \draw[draw=black, ultra thick, miter limit=10.0, -stealth]
        (peg.center) -- node[pos=0, xshift=0.5*\pegwidth\textwidth + 0.25*\tablesidewidth\textwidth, yshift=-0.125*\pegheight\textwidth, font=\small]{push}
        (\tableslotcenterx, \tableslotcentery);
    
    \path[draw=black,fill=my-light-orange,very thick,miter limit=10.0]
        (\tablesidewidth, \tableheight) --
        (\tablesidewidth, \tableslotheight) --
        (\tablewidth-\tablesidewidth, \tableslotheight) --
        (\tablewidth-\tablesidewidth, \tableheight) --
        (\tablewidth, \tableheight) --
        (\tablewidth, 0) --
        (0, 0) --
        (0, \tableheight) --
        cycle;
\end{tikzpicture}
\vfill
        \caption{Insertion}
        \label{fig:peg-in-hole-insertion}
    \end{subfigure}
    
    \caption{Peg-in-hole task}
    \label{fig:peg-in-hole-phases}
\end{figure*}

In literature, PiH is considered as a benchmark for force control strategies applied to contact-rich assembly tasks \cite{jiang_state---art_2020}.
As displayed in Figure~\ref{fig:peg-in-hole-phases}, it consists of different phases: first, the robot searches for the hole location where the held peg has to be inserted, typically with translational movements (Figure~\ref{fig:peg-in-hole-search}).
When the peg engages the hole, the EE rotates to align the former against the latter's walls (Figure~\ref{fig:peg-in-hole-engagement}).
Lastly, the insertion phase actually places the peg into the hole slot (Figure~\ref{fig:peg-in-hole-insertion}).

For this task, inherent difficulties arise, caused by sub-millimetric tolerance between peg's and hole's dimensions, inaccuracies in estimating the hole's exact location, and complex forces and torques to be managed at the EE.
In recent years, RL has been the most popular approach with which these challenges have been faced \cite{ji_deep_2024}, as it usually does not rely on a specific model, instead trying to optimize motion policies according to a tailored reward function, which is constantly updated as data are collected during the task execution.

\subsection{Stability and safety}

The first practical objective RL usually struggles to accomplish is \textit{guaranteed asymptotic stability}, \ie it does not usually provide a theoretical proof formally guaranteeing the RL policy to actually converge towards the objective.
In \cite{khader_stability-guaranteed_2021}, this paramount problem is analyzed, thus devising RL policies with guaranteed stability.
To achieve this feature, a \textit{variable} impedance controller\footnote{A variable impedance controller is similar to the one in \eqref{eq:impedance-controller}, where $K_p$ and/or $K_d$ are updated online in an outer optimization loop.} -- originally defined in \cite{khansari_modeling_2014} -- is proposed, which is globally asymptotically stable if its matrices are symmetric positive definite (SPD): so, \cite{khader_stability-guaranteed_2021} proposes to generate them from Wishart distributions.
Interestingly, in the PiH experiment, \cite{khader_stability-guaranteed_2021} clarifies that the stiffness and damping matrices of the variable impedance controller defined in \cite{khansari_modeling_2014} are initialized as, \textit{but not constrained to}, diagonal matrices, hence non-diagonal matrices are computed by the RL policy when solving the task.
This is in contrast with classical methods, as $K_p$ and $K_d$ in \eqref{eq:impedance-controller} are usually chosen as diagonal matrices.

\begin{figure}[b!]
    \centering
    \begin{tikzpicture}[auto, font=\footnotesize]
        \definecolor{yellow}{HTML}{FFF2CC}
        \definecolor{darkgreen}{HTML}{82B366}
        \definecolor{lightgreen}{HTML}{D5E8D4}
        \definecolor{darkred}{HTML}{B85450}
        \definecolor{lightred}{HTML}{F8CECC}
    
        \pgfmathsetmacro{\width}{3/11*\columnwidth}
        \pgfmathsetmacro{\height}{1/11*\columnwidth}
        \pgfmathsetmacro{\hdist}{1/11*\columnwidth}
        \pgfmathsetmacro{\vdist}{2/11*\columnwidth}
        
        \tikzstyle{block} = [rectangle, rounded corners, minimum width=\width, minimum height=\height, text centered, thick, draw=black, fill=yellow]
        \tikzstyle{offline} = [block, fill=lightred, draw=darkred]
        \tikzstyle{online} = [block, fill=lightgreen, draw=darkgreen]
        \tikzstyle{diamondselector} = [diamond, minimum height=\height, text centered, thick, draw=black, fill=yellow, aspect=2]
    
        \tikzstyle{line} = [thick, draw=black]
        \tikzstyle{arrow} = [-stealth, line]
        \tikzstyle{redarrow} = [arrow, draw=darkred]
        \tikzstyle{greenarrow} = [arrow, draw=darkgreen]
        
        \node (offline) [offline] {Offline data};
        \node (safety) [block, right of=offline, node distance=\width+\hdist] {Safety Critic};
        \node (online) [online, right of=safety, node distance=\width+\hdist] {Online data};
        \node (selector) [diamondselector, below of=safety, node distance=\vdist] {$\varepsilon > \alpha$};
        \node (recovery) [block, below of=selector, node distance=\vdist, xshift=-2*\hdist] {Recovery Policy};
        \node (task) [block, below of=selector, node distance=\vdist, xshift=2*\hdist] {Task Policy};
        \node (formula) [block, below of=selector, node distance=2*\vdist] {$f_c = K_p \tilde x + K_d \dot x$};
    
        \coordinate (general_point) at ([yshift=0.25*\vdist]formula.north);
    
        \node[circle, fill=black, minimum size=4pt, inner sep=0pt] at (general_point) {};
    
        \draw[redarrow] (offline) -- (safety);
        \draw[redarrow] ([xshift=-0.25*\width]offline.south) |- (recovery.west);
        \draw[greenarrow] (online) -- (safety);
        \draw[greenarrow] (online.south) |- (task.east);
        \path[greenarrow]
            ([xshift=0.25*\width]online.south) |-
            ([xshift=0.125*\width, yshift=-0.25*\vdist]task.south west) --
            (recovery.east);
        \draw[arrow] (safety) -- node[left] {$\varepsilon$} (selector);
        \draw[arrow] (selector.west) -| node[left] {Y} (recovery.north);
        \draw[arrow] (selector.east) -| node[right] {N} (task.north);
        \path[line]
            (recovery.south) --
            ([yshift=-0.125*\vdist]recovery.south) --
            (general_point);
        \draw[line]
            (task.south) --
            ([yshift=-0.125*\vdist]task.south) --
            (general_point);
        \draw[arrow] (general_point) --node[anchor=east, font=\scriptsize]{$K_p, x_d$} (formula.north);
    \end{tikzpicture}
    \caption{Safe Reinforcement Learning with Variable Impedance Control (SRL-VIC) used in \cite{zhang_srl-vic_2024}}
    \label{fig:srl-vic}
\end{figure}
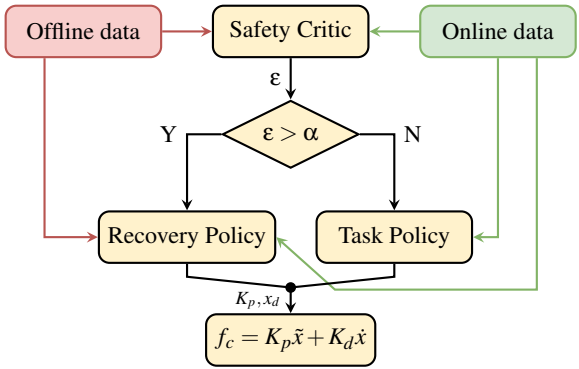

The same issue is considered with another approach in \cite{zhang_srl-vic_2024}, which proposes a variable-stiffness impedance controller, where the stiffness matrix $K_p$ and the task-space reference $x_d$ in \eqref{eq:impedance-controller} are computed with two RL policies. 
This control scheme is applied to the contact-rich cable routing manipulation task, where the robot has to explore a path constrained by walls, whose location is unknown, hence the robot can only rely on contact forces to have a feedback on the environment.
The two policies are called ``task policy'' and ``recovery policy'': the former is used to accomplish the task (\ie, reaching the goal at the end of the maze), and the latter is used to recover from an \textit{unsafe} state-action pair (see Figure~\ref{fig:srl-vic}).
It is noteworthy to specify that stability is not formally guaranteed, but safety is promoted with the concept of ``risk learning'': indeed, a ``safety critic'' network is trained to compute the ``degree of risk'' $\epsilon$ of a given task policy action.
If $\epsilon$ is above a safety threshold $\alpha$, then the recovery policy action is applied.
The ``risk learning'' phase is performed in a preliminary offline training, where the risk (\ie, the output of the ``safety critic'' network) depends on the satisfaction of a safety constraint requiring the measured force to stay below a predefined threshold.

In order to stress how the aspect of safety is a strict requirement in these tasks, it is worth pointing out that, whenever it is not explicitly solved with tailored strategies, it is common for researchers to clip the RL policy action in a limited bound \cite{pozzi_experimental_2023}.
For instance, this guideline is followed by \cite{hou_fuzzy_2022}, which performs the multiple PiH task updating stiffness and damping parameters with a fuzzy-logic controller, and selecting the optimal EE control action making use of DQN \cite{mnih_playing_2013} and DDPG \cite{lillicrap_continuous_2016} RL frameworks, limiting the action of the latter in a bounded range.

\subsection{Optimization strategies}

Although PiH is one of the most common assembly tasks, there is still no consolidated methodology to approach it.
Indeed, even considering the spectrum of RL-based techniques, various diverse strategies have been proposed, both in terms of \textit{actions} (\ie, the output of the RL policy) and optimization algorithms used to devise the policy itself.

Indeed, the already referenced works choose the action semantics to be either $K_p$ \cite{khader_stability-guaranteed_2021} or $x_d$ \cite{hou_fuzzy_2022} or both \cite{zhang_srl-vic_2024}.
However, other approaches are possible: for instance, \cite{ji_deep_2024} solves the task by separating the phases of hole searching and hole insertion (see Figure~\ref{fig:peg-in-hole-phases}) and optimizing two different configurations of \textit{non-diagonal} (similarly to \cite{khader_stability-guaranteed_2021}) proportional matrices $K_P$ in \eqref{eq:direct-force-control} through DDPG.

All the aforementioned works rely on model-free frameworks.
Instead, \cite{luo_reinforcement_2019} tackles an assembly task with a model-based RL approach, \ie, iLQG \cite{todorov_generalized_2005}, with which it is possible to compute an optimal policy in closed form, outputting the impedance Cartesian wrench $f_c$ directly.

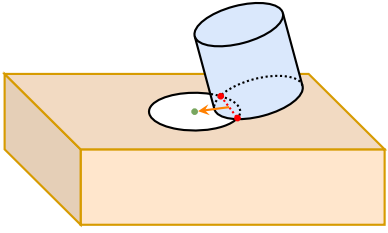
\begin{figure}[t!]
    \centering

    \tikzset{
        partial ellipse/.style args={#1:#2:#3}{
            insert path={+ (#1:#3) arc (#1:#2:#3)}
        }
    }
    
    \begin{tikzpicture}
        \def\tableWidth{4}
        \def\tableHeight{1}
        \def\tableDepth{1}
    
        \def\holeXc{0.5*\tableWidth - 0.5*\tableDepth}
        \def\holeYc{\tableHeight + 0.5*\tableDepth}
        \def\holeXr{0.15*\tableWidth}
        \def\holeYr{0.25*\tableDepth}
    
        \def\rotationAngle{15}
        \def\startAngle{69}
        \def\endAngle{20}
        \def\contactAngleA{55}
        \def\contactAngleB{-\endAngle}
        
        \def\pegHeight{\tableHeight}
        \def\pegXr{\holeXr}
        \def\pegYr{\holeYr}
        \def\pegXc{\holeXc + 0.7*\holeXr + 0.7*\pegXr}
        \def\pegYc{\holeYc + 0.75*\pegYr}
    
        \def\circleRadius{0.03}
        \pgfmathsetmacro{\contactPointAX}{\holeXc + \holeXr * cos(\contactAngleA)}
        \pgfmathsetmacro{\contactPointAY}{\holeYc + \holeYr * sin(\contactAngleA)}
        \pgfmathsetmacro{\contactPointBX}{\holeXc + \holeXr * cos(\contactAngleB)}
        \pgfmathsetmacro{\contactPointBY}{\holeYc + \holeYr * sin(\contactAngleB)}
        \pgfmathsetmacro{\contactPointMX}{(\contactPointAX + \contactPointBX) / 2}
        \pgfmathsetmacro{\contactPointMY}{(\contactPointAY + \contactPointBY) / 2}
        
        \definecolor{frontColor}{HTML}{FFE6CC}
        \definecolor{sideColor}{HTML}{E5CFB7}
        \definecolor{topColor}{HTML}{F2DAC2}
        \definecolor{lineColor}{HTML}{D79B00}
        \definecolor{cylinderColor}{HTML}{DAE8FC}
        \definecolor{holeCenterColor}{HTML}{78A65F}
        \definecolor{contactPointColor}{HTML}{FF0000}
    
        \draw[fill=frontColor, draw=lineColor, thick] (0, 0) rectangle (\tableWidth, \tableHeight);
        \draw[fill=sideColor, draw=lineColor, thick] (0, 0) -- ++(-\tableDepth, \tableDepth) -- ++(0, \tableHeight) -- ++(\tableDepth, -\tableDepth) -- cycle;
        \draw[fill=topColor, draw=lineColor, thick] (0, \tableHeight) -- ++(-\tableDepth, \tableDepth) -- ++(\tableWidth, 0) -- ++(\tableDepth, -\tableDepth) -- cycle;
    
        \fill[white] (\holeXc, \holeYc) ellipse [x radius=\holeXr, y radius=\holeYr];
        \draw[thick, black]
            (\holeXc, \holeYc)
            [partial ellipse=\startAngle:360-\endAngle:{\holeXr} and {\holeYr}];
        
        \begin{scope}[rotate around={\rotationAngle:(\pegXc, \pegYc)}]
            \fill[cylinderColor] 
                (\pegXc - \pegXr, \pegYc) -- 
                ++(0, \pegHeight) --
                ++(2*\pegXr, 0) --
                ++(0, -\pegHeight) --
                cycle;
    
            \fill[cylinderColor, draw=black, thick] (\pegXc, \pegYc + \pegHeight) ellipse [x radius=\pegXr, y radius=\pegYr];
    
            \fill[cylinderColor] (\pegXc, \pegYc) ellipse [x radius=\pegXr, y radius=\pegYr];
    
            \draw[draw=black, thick]
                (\pegXc + \pegXr, \pegYc)
                arc [start angle=0, end angle=-180, x radius=\pegXr, y radius=\pegYr];
            \draw[densely dotted, draw=black, thick]
                (\pegXc + \pegXr, \pegYc)
                arc [start angle=0, end angle=180, x radius=\pegXr, y radius=\pegYr];
    
            \draw[draw=black, thick]
                (\pegXc - \pegXr, \pegYc) --
                ++(0, \pegHeight);
            \draw[draw=black, thick]
                (\pegXc + \pegXr, \pegYc) --
                ++(0, \pegHeight);
        \end{scope}
    
        \draw[densely dotted, thick, black]
            (\holeXc, \holeYc)
            [partial ellipse=\startAngle:-\endAngle:{\holeXr} and {\holeYr}];
    
        \draw[fill=contactPointColor, draw=contactPointColor, thick] (\contactPointAX, \contactPointAY) circle [radius=\circleRadius];
        \draw[fill=contactPointColor, draw=contactPointColor, thick] (\contactPointBX, \contactPointBY) circle [radius=\circleRadius];
    
        \draw[contactPointColor, thick, densely dotted] (\contactPointAX, \contactPointAY) -- (\contactPointBX, \contactPointBY);
    
        \pgfmathsetmacro{\goalDirX}{\holeXc - \contactPointMX}
        \pgfmathsetmacro{\goalDirY}{\holeYc - \contactPointMY}    
        \pgfmathsetmacro{\goalDistance}{sqrt(\goalDirX*\goalDirX + \goalDirY*\goalDirY)}
        \pgfmathsetmacro{\shortenFactor}{0.9} 
        \pgfmathsetmacro{\shortenedGoalX}{\contactPointMX + (\goalDirX * \shortenFactor)}
        \pgfmathsetmacro{\shortenedGoalY}{\contactPointMY + (\goalDirY * \shortenFactor)}    
        \draw[-stealth, orange, thick] (\contactPointMX, \contactPointMY) -- (\shortenedGoalX, \shortenedGoalY);
    
        \draw[fill=holeCenterColor, draw=holeCenterColor, thick] (\holeXc, \holeYc) circle [radius=\circleRadius];
        
    \end{tikzpicture}

    \caption{PiH approach used in \cite{unten_peg--hole_2023}: the contact points (in red) are estimated, and a translational motion (in orange) is planned towards the goal (in green)}
    \label{fig:peg-in-hole-geometric}
\end{figure}

Lastly, it is worth mentioning a recent novel approach that actually falls outside the RL realm.
In fact, \cite{unten_peg--hole_2023} exploits force-related information to devise a ``motion planning'' approach to effectively solve the PiH problem.
In particular, given force/torque measurements and known peg geometry, the contact points at which the peg enters in contact with the hole walls are estimated.
Then, from the two contact points the direction along which the peg must be moved to precisely match the hole is computed as the line connecting the midpoint and the hole center (see Figure~\ref{fig:peg-in-hole-geometric}).
Currently, it has never been assessed whether this solution is advantageous compared to RL: as will be discussed in Section~\ref{sec:future-directions}, we believe this aspect is a prospect to be analyzed in future research.

\subsection{Reward formulation}\label{sec:reward-formulation}

As mentioned in Section~\ref{sec:problem-definition}, formulating a significant reward is fundamental for a RL approach to succeed.
Indeed, the reward function $R$ should coherently express the goal of the task, so as to drive the optimization algorithm to yield the optimal policy.

In RL-based PiH, the most popular rewards are Euclidean distance to goal position $x_g$, in the form
\begin{equation}\label{eq:reward-distance}
    R_g = -\lVert x_g - x \rVert^2,
\end{equation}
and the time to complete the task, expressed as
\begin{equation}\label{eq:reward-time}
    R_T = 1 - \frac{k}{T},
\end{equation}
where $k \in \mathbb N$ is the current time step and $T \in \mathbb N$ a parameter denoting the maximum number of steps the task should be completed in.

Additionally, some penalties may be added for the sake of safety, \eg, a penalty on the norm of the action $a$:
\begin{equation}\label{eq:penalty-action}
    R_a = -\lVert a \rVert^2,
\end{equation}
or a safety penalty to avoid the manipulator to exert excessive forces, \ie,
\begin{equation}\label{eq:penalty-force}
    R_f =
    \begin{cases}
        -P, & f_e > f_{th}\\
        0, & \text{otherwise}
    \end{cases},
\end{equation}
where $f_{th} \in \mathbb R^+$ is the safety threshold and $P \in \mathbb R^+$ is the penalty.

\begin{figure}[t!]
    \centering

    \begin{subfigure}{0.32\columnwidth}
        \centering
        \includegraphics[width=\textwidth]{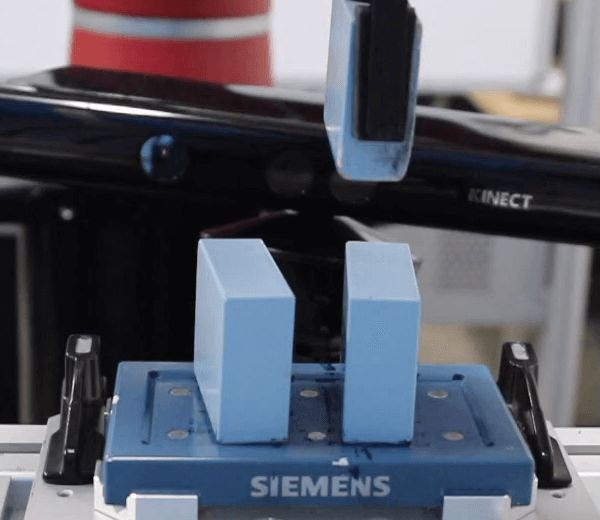}
    \end{subfigure}
    \hfill
    \begin{subfigure}{0.32\columnwidth}
        \centering
        \includegraphics[width=\textwidth]{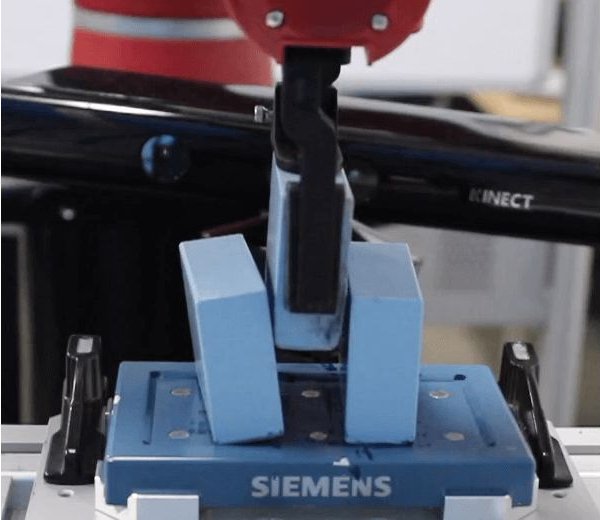}
    \end{subfigure}
    \hfill
    \begin{subfigure}{0.32\columnwidth}
        \centering
        \includegraphics[width=\textwidth]{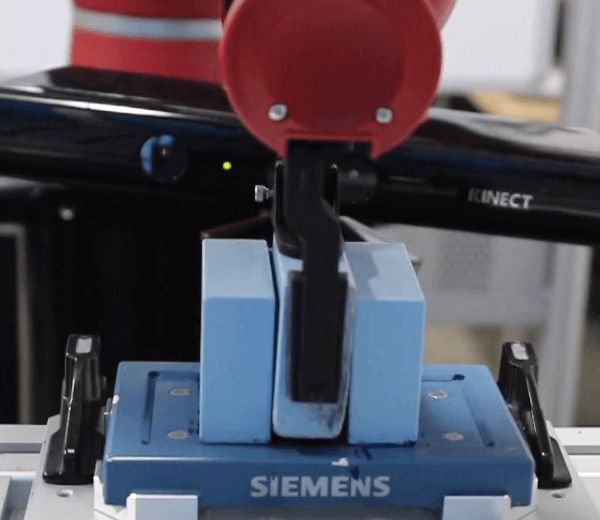}
    \end{subfigure}
    
    \caption{Residual RL approach \cite{johannink_residual_2019}}
    \label{fig:rrl}
\end{figure}

A unique reward is used by \cite{johannink_residual_2019}, in which the RL action $u_r$ is added as a residual to that of a low-level controller: in particular, with the objective of inserting a peg in a hole between two fixed blocks (see Figure~\ref{fig:rrl}), the reward includes a contribution to describe how much the ``left'' and ``right'' blocks are tilting from their upright position:
\begin{equation}
    R_h = - \alpha_\theta \left( \lvert \theta_l \rvert + \lvert \theta_r \rvert \right) - \alpha_\phi \left( \lvert \phi_l \rvert + \lvert \phi_r \rvert \right),
\end{equation}
where $\theta$ and $\phi$ represent the hole blocks' tilting angles, and $\alpha_\theta, \alpha_\phi \in \mathbb R^+$ are hyperparameters.

As evident, the reward does not always follow a particular formulation: researchers typically tend to design them according to certain safety, accuracy or time requirements.
In the next section, we will report what the reference works choose as reward, possibly linearly combining them in the form (\eg summing \eqref{eq:reward-distance} and \eqref{eq:penalty-action})
\begin{equation}
    R = \alpha_g R_g + \alpha_a R_a,    
\end{equation}
with $\alpha_g, \alpha_a \in \mathbb R^+$ being two weights.

\begin{table*}[t]
\centering
\caption{Differences between PiH approaches. Reward functions sums are intended to be weighted sums.}
\label{tab:rewards}
\begin{tabular}{c|c|c|c|c|c|c}
\textbf{Reference} & \textbf{Reward} & \textbf{Stability} & \textbf{RL algorithm} & \textbf{Actions} & \textbf{Unified} & \textbf{Simulator} \\ 
\hline
\cite{khader_stability-guaranteed_2021} & $R_g + R_a$ & \cmark & CEM-like & $K_p, K_d$ & \cmark & MuJoCo \\
\cite{ji_deep_2024} & $R_f + R_T$ & \xmark & DDPG & $K_P$ & \xmark & PyBullet \\
\cite{narang_factory_2022} & $R_g$ & \xmark & PPO & $x_d$ & \cmark & IsaacGym \\
\cite{hou_fuzzy_2022} & $R_f + R_T$ & \xmark & DQN, DDPG & $x_d$ & \cmark & --- \\
\cite{zhang_srl-vic_2024} & $R_g + R_f$ & \cmark & DDPG & $K_p, x_d$ & \cmark & MuJoCo \\
\cite{luo_reinforcement_2019} & $R_g$ & \xmark & iLQG & $f_c$ & \cmark & --- \\
\cite{tang_industreal_2023} & $R_g$ & \xmark & PPO & $x_d$ & \cmark & IsaacGym \\
\cite{johannink_residual_2019} & $R_g + R_h$ & \xmark & TD3 & $u_r$ & \cmark & MuJoCo
\end{tabular}
\end{table*}

\subsection{Force-tracking tasks}\label{sec:force-tracking-tasks}

Although PiH does not require force tracking, similar strategies and learning frameworks have been applied to pursue this specific requirement as well, as mentioned in Section~\ref{sec:introduction}.
Such methodologies are developed with the aim of increasing the performance of a standard PI DFC \eqref{eq:direct-force-control} in terms of force-tracking error $\tilde f$.
For instance, \cite{pozzi_experimental_2023} uses RL to compute the optimal setpoint $x_d$ for an impedance controller, minimizing $\tilde f$.

Other relevant works employing AI foresee exploiting Neural Networks (NNs) in learning interaction dynamics in variable damping \cite{huang_optimal_2021,hamedani_intelligent_2021} and variable stiffness \cite{liu_optimized_2021,anand_model-based_2023} impedance controllers, or selecting the optimal action enhancing that of a low-level DFC \cite{petrone_optimized_2024}.
It is evident that, in fact, similar AI-enhanced force controllers are employable in various contexts, possibly different than that of PiH.

Nevertheless, in industrial contexts, solely ensuring force-tracking accuracy might not be completely satisfactory: indeed, AI-based methods are not currently coping with practical problems such as tool wear \cite{lloyd_precision_2024} in deburring applications (Figure~\ref{fig:deburring}) and high-speed motions \cite{iskandar_hybrid_2023} in polishing-like tasks (Figure~\ref{fig:polishing}).
We claim that both these aspects are of utmost importance in industries, since they aim at maximizing the efficiency, efficacy and quality of the delivered products, thus a future challenge for AI in force control is properly managing these problems, in conjunction with force-tracking.

\subsection{Summary}

In Table~\ref{tab:rewards} we summarize the major differences in the PiH approaches discussed in the previous sections, in terms of
\begin{enumerate*}[label=(\roman*)]
    \item reward;
    \item explicit or implicit stability;
    \item RL algorithm\footnote{More details on PPO and TD3 can be found in \cite{schulman_proximal_2017} and \cite{fujimoto_addressing_2018}};
    \item RL policy action;
    \item unification of the hole-searching and hole-insertion phases;
    \item simulator.
\end{enumerate*}
Simulators are paramount in rapidly training RL policies, but their fidelity \wrt crucial factors such as system dynamics and realism in force/torque measures may heavily influence their success when deployed on the real hardware \cite{sorensen_online_2016}\footnote{Popular examples are, as also listed in Table~1, MuJoCo \cite{todorov_mujoco_2012}, IsaacGym \cite{makoviychuk_isaac_2021} and PyBullet \cite{coumans_pybullet_2016}}.
In this sense, \cite{narang_factory_2022} proposed a realistic dataset of simulated assets, and \cite{tang_industreal_2023} introduced specific learning and control strategies to bridge the gap between simulated and real worlds.

Given the evident diversities among recently proposed literature in all the major aspects of the selected approaches, we stress that facing the peculiar PiH task with RL can still be considered an open problem, as a consolidated solution does not currently exist.
Our position is that future research should concentrate on limiting the gaps between these strategies, formally comparing their peculiarities or possibly merging their advancements.
We will further elaborate this claim in Section~\ref{sec:future-directions}.

\section{\uppercase{Future Directions}}\label{sec:future-directions}

In the light of the discussions done in Section~\ref{sec:challenges}, we now state our position on the topic of empowering manufacturing processes with AI methods.
We deem that future research on this subject should concentrate on consolidating the novel technologies that are rapidly emerging in the latest years, both in force-tracking applications and in contact-rich assembly tasks.

To this aim, we suggest to devise common methods to formally compare RL-based techniques, both among them and against standard approaches, defining quantitative metrics according to which existing and novel methodologies shall be validated.
For instance, possible performance metrics in PiH might be
\begin{enumerate*}[label=(\roman*)]
    \item success rate;
    \item amount of exerted forces on the workpiece;
    \item execution time.
\end{enumerate*}
In this sense, it is required to define a specific reward shape (see Section~\ref{sec:reward-formulation}), and to formally state performance requirements in terms of, \eg, peg-hole tolerance, so as to fairly compare these methods.

We consider some of the features of the referenced works to be essential in practical scenarios, \eg guaranteed asymptotic stability in \cite{khader_stability-guaranteed_2021}.
Hence, we deem that, in the future, researchers shall make an effort in trying to integrate the various enhancements independently proposed in the relevant recent literature, in order to accomplish various performance requirements, as stated above.

As regards other relevant applications we discussed, namely deburring and polishing, we claim that researchers and practitioners should continue to pursue methods minimizing force-tracking error, as it clearly is the most immediate quantitative index describing the quality of such tasks.
However, it is of sheer importance to ensure that novel methods deal with the subjects highlighted in Section~\ref{sec:force-tracking-tasks}, \ie performance degradation in highly dynamic scenarios and impact on workpiece machining quality and tool wear.
To the best of the authors' knowledge, such topics are not currently covered by AI-driven methods, thus we suggest to invest on this direction, in order to increase the relevance of future works in both academic and industrial contexts.

These considerations are in line with the final goal of providing equivalence with standard and consolidated approaches in terms of \textit{perceived compatibility}.
This aspect is indeed a paramount enabler for the use of AI in manufacturing, as demonstrated in \cite{merhi_enablers_2023}, according to which ``any innovation is considered compatible with an organization only when it is perceived as consistent with existing business processes, practices, and values''.

\section{\uppercase{Conclusions}}\label{sec:conclusions}

This paper reported recent advancements on AI methodologies applied to manufacturing robotic tasks.
The rationale behind employing these technologies is two-fold.
First, in the context of Industry 4.0, they can further optimize manufacturing processes, increasing the production quality, efficiency and throughput.
Moreover, they can compensate for inherent limits of classical model-based control methods, as usually happens in challenging force-related and contact-rich applications.

We analyzed issues and objectives these methods are demanded to undertake when applied in real-world industrial scenarios, and analyzed the differences in recent relevant research works on this topic, both on methodological and implementation-related aspects.
In conclusion, we claimed our position on possible directions future research should pursue, in order to accommodate for specific performance requirements, and proposed suggestions researchers shall potentially follow to increase the relevance of future works on both academic and practical level.

\bibliographystyle{apalike}
{\small
\bibliography{IEEEabrv, OtherAbbrv, icinco-2024-ai-force-control}}

\end{document}